\documentclass[letterpaper]{article} 
\usepackage{aaai2026}  
\usepackage{times}  
\usepackage{helvet}  
\usepackage{courier}  
\usepackage[hyphens]{url}  
\usepackage{graphicx} 
\usepackage{natbib}  
\usepackage{caption} 
\usepackage{amssymb}   
\usepackage{amsmath}  

\usepackage{algorithmic}
\usepackage{newfloat}
\usepackage{listings}
\DeclareCaptionStyle{ruled}{labelfont=normalfont,labelsep=colon,strut=off} 
\title{Steering Sparse Autoencoder Latents to Control Dynamic Head Pruning in \\ Vision
Transformers (Student Abstract)}
\author{
    Yousung Lee\textsuperscript{\rm 1}, 
    Dongsoo Har\textsuperscript{\rm 1}
}
\affiliations{
    \textsuperscript{\rm 1}Korea Advanced Institute of Science and Technology, Daejeon 34051, Korea \\
    yslee410@kaist.ac.kr, dshar@kaist.ac.kr
}

\usepackage{bibentry}

\begin{document}

\maketitle

\begin{abstract}

Dynamic head pruning in Vision Transformers (ViTs) improves efficiency by removing redundant attention heads, but existing pruning policies are often difficult to interpret and control. In this work, we propose a novel framework by integrating Sparse Autoencoders (SAEs) with dynamic pruning, leveraging their ability to disentangle dense embeddings into interpretable and controllable sparse latents. Specifically, we train an SAE on the final-layer residual embedding of the ViT and amplify the sparse latents with different strategies to alter pruning decisions. Among them, per-class steering reveals compact, class-specific head subsets that preserve accuracy. For example, \textit{bowl} improves accuracy (76\%$\rightarrow$82\%) while reducing head usage (0.72$\rightarrow$0.33) via heads $h_2$ and $h_5$. These results show that sparse latent features enable class-specific control of dynamic pruning, effectively bridging pruning efficiency and mechanistic interpretability in ViTs.

\end{abstract}

\section{Introduction}

Vision Transformers (ViTs) leverage multi-head self-attention to capture diverse token interactions. 
However, many attention heads are redundant, increasing computation without proportional performance gain. 
To address this, adaptive frameworks such as AdaViT \cite{meng2022adavit} have been introduced, which use auxiliary networks to select which heads to prune.
This input-dependent pruning substantially reduces computation while preserving accuracy.

However, a key limitation remains: since these pruning policies rely on residual embeddings, the decision process is often opaque and difficult to control at the latent level. As a result, while dynamic head pruning improves efficiency, its mechanism lacks interpretability. If pruning decisions could be explained or controlled at the latent level, head selection would become both interpretable and controllable. 

Sparse Autoencoders (SAEs) offer a natural tool for this purpose, as they disentangle dense, polysemantic embeddings into sparse latent features that tend to encode more monosemantic and interpretable concepts in transformer representations \cite{cunningham2023sparse, lim2025patchsae}. Recent studies have shown that this disentanglement enables steering specific SAE latent dimensions to control model behavior in a desired direction \cite{kang2024interpret, chatzoudis2025vs2}.

\begin{figure}[t]
    \centering
    \includegraphics[width=1.0\columnwidth]{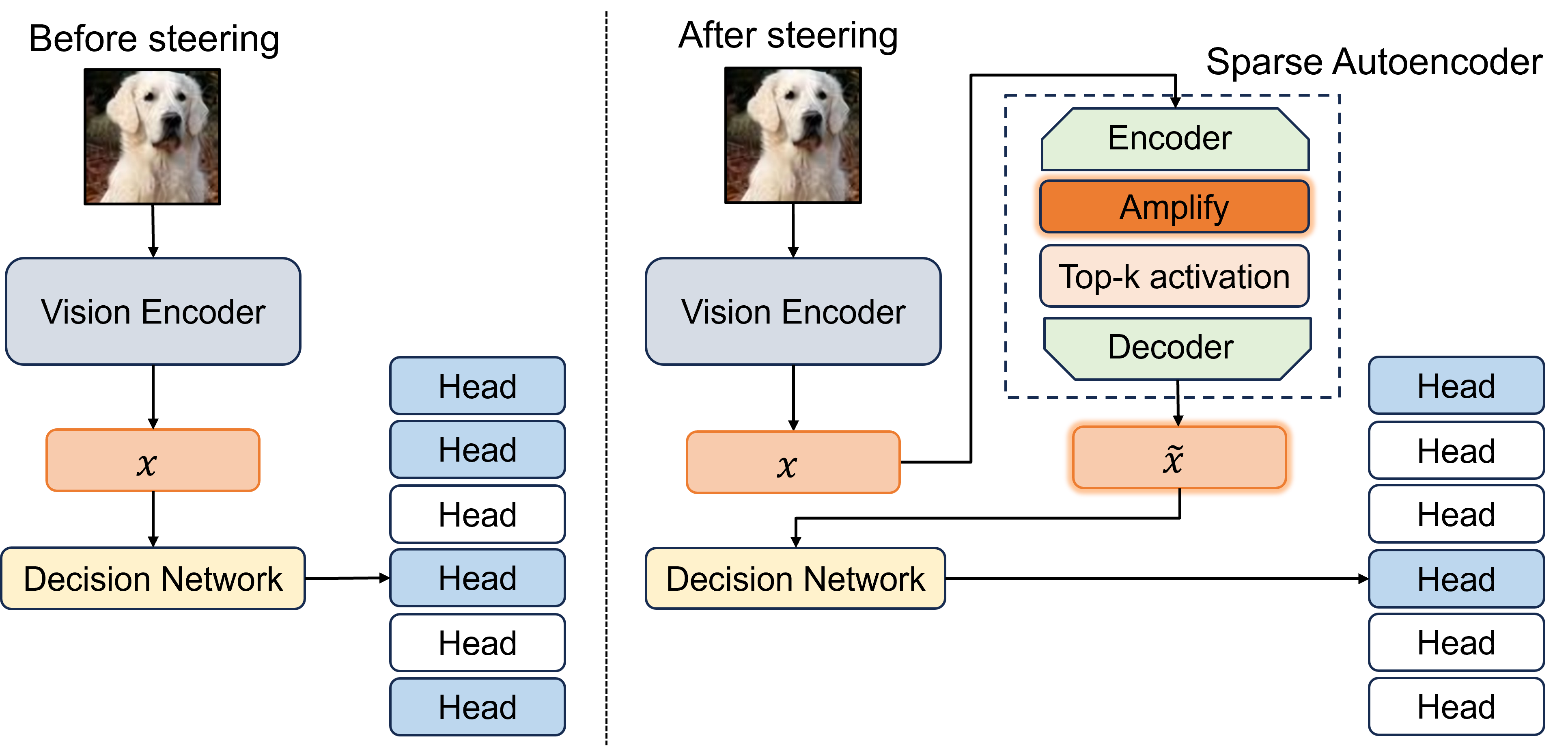} 
    \caption{Comparison of dynamic head pruning before (left) and after (right) SAE latent steering, where $x$ denotes the \texttt{CLS} token from the final-layer residual stream input, and $\tilde{x}$ denotes the steered embedding.}
    \label{fig1}
    \end{figure}

In this work, we propose a novel framework that integrates $k$-sparse autoencoders \cite{makhzani2013k} with dynamic head pruning in ViTs to make pruning decisions controllable at the latent level.
This framework also enhances interpretability by revealing class-specific head subsets.
As illustrated in Figure~\ref{fig1}, we amplify selected SAE latent dimensions, reconstruct the steered embedding, and feed it into the decision network to observe how pruning behavior changes. Our experiments show that per-class latent steering is particularly effective, reducing head usage while largely maintaining accuracy. Overall, these results suggest that sparse latents provide an effective way to interpret and control dynamic pruning, bridging the gap between efficiency and mechanistic interpretability in ViTs.

\section{Method}
    
In this work, we adopt AdaViT as the baseline dynamic pruning framework, with a particular focus on head pruning. 
We first train a Vision Transformer with layer-wise decision networks. In this setup, each lightweight network receives the class (\texttt{CLS}) token from the residual stream input and outputs head importance logits $a_\ell \in \mathbb{R}^H$, where $\ell$ denotes the layer index and $H$ the number of attention heads.
We use the \texttt{CLS} token as input for the decision network since it encodes global context relevant for classification.
Binary masks $M_{\ell,i} \in \{0,1\}$ are obtained via Gumbel-Sigmoid sampling from $a_{\ell,i}$, 
where $i$ denotes the head index. The masked attention is computed as follows:
\begin{equation}
h_{\ell,i} = M_{\ell,i}\,\mathrm{Attn}(Q,K,V)_{\ell,i}.
\end{equation}

The Vision Transformer and the decision network are trained jointly to preserve accuracy while enforcing head sparsity toward a target head usage ratio. 
After training, we extract $x \in \mathbb{R}^d$, 
the \texttt{CLS} token from the final layer’s residual input and use it to train a $k$-sparse autoencoder.
Formally, the encoder and decoder are given by
\begin{align}
z &= \mathrm{TopK}\!\left(W_{\mathrm{enc}}(x - b_{\mathrm{dec}})\right), 
\quad W_{\mathrm{enc}} \in \mathbb{R}^{n \times d}, \\
\hat{x} &= W_{\mathrm{dec}} z + b_{\mathrm{dec}}, 
\quad W_{\mathrm{dec}} \in \mathbb{R}^{d \times n},
\end{align}
where $z$ is the sparse latent representation of $x$, $\hat{x}$ is its reconstruction, 
and $b_{\mathrm{dec}}$ denotes the decoder bias.
The parameters $n$ and $d$ represent the SAE latent and input embedding dimensions, respectively. 
Sparsity is enforced through a top-$k$ activation, which preserves only the $k$ largest dimensions. 
The SAE is trained with the following mean squared error (MSE) reconstruction objective:
\begin{equation}
\mathcal{L}_{\mathrm{rec}} = \|x - \hat{x}\|_2^2,
\end{equation}
which encourages $\hat{x}$ to remain close to $x$.
After training the SAE, we amplify the selected latent dimensions $S$ by
\begin{equation}
z_i' =
\begin{cases}
z_i + \alpha, & i \in S,\\[3pt]
z_i, & i \notin S,
\end{cases}
\quad
\tilde{z} = \mathrm{TopK}(z'),
\end{equation}
where $\alpha$ is the amplification strength and $\tilde{z}$ denotes the 
steered sparse latent vector obtained after amplification and top-$k$ activation.
The steered embedding is then reconstructed as
$\tilde{x} = W_{\text{dec}}\tilde{z} + b_{\text{dec}}$,
which is finally fed into the decision network to obtain the steered pruning mask.

\section{Experiments}
\paragraph{Dynamic pruning baseline.}
We fine-tune an ImageNet-pretrained ViT-Small ($12$ layers, 
$6$ heads per layer, $384$-d embeddings) on CIFAR-$100$, 
reaching $91.27\%$ accuracy. 
Each layer employs a single decision network for head selection, as described in the Method section. 
The final model prunes $30\%$ of heads while maintaining $89.79\%$ accuracy.

\paragraph{Sparse autoencoder.}
The SAE expands the $384$-d residual embedding into the $3072$-d latent space ($8\times$) with top-$k$ activation ($k=64$).
It is trained for $100$ epochs, achieving an MSE loss of $0.0228$. 
Replacing the original embedding with its reconstruction yields only minor differences in accuracy ($-0.12\%$) and head usage ratio ($+0.025$), showing that the SAE preserves essential information for effective pruning.

\begin{figure}[t]
    \centering
    \includegraphics[width=1.0\columnwidth]{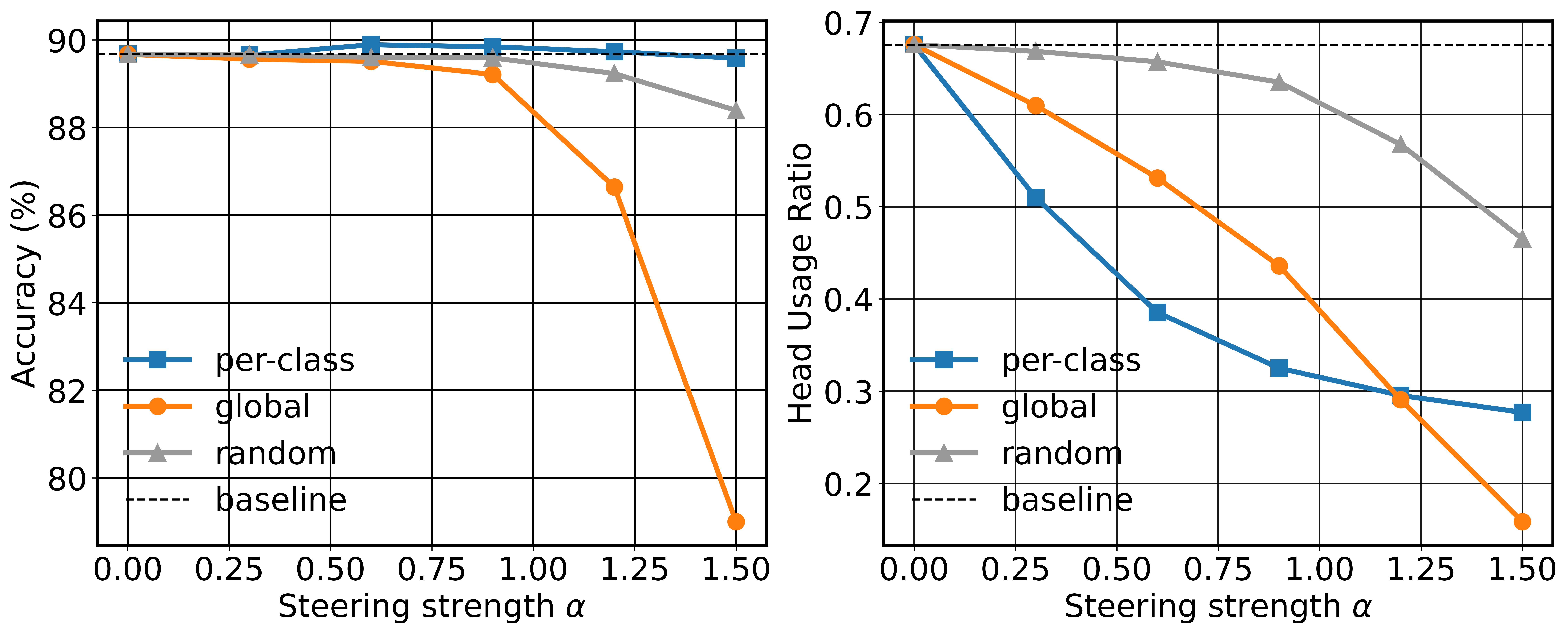}
    \caption{Accuracy (\%) and head usage ratio in the final layer under different strategies as $\alpha$ increases from $0$ to $1.5$.}
    \label{fig2}
\end{figure}

\begin{figure}[t]
    \centering
    \includegraphics[width=1.0\columnwidth]{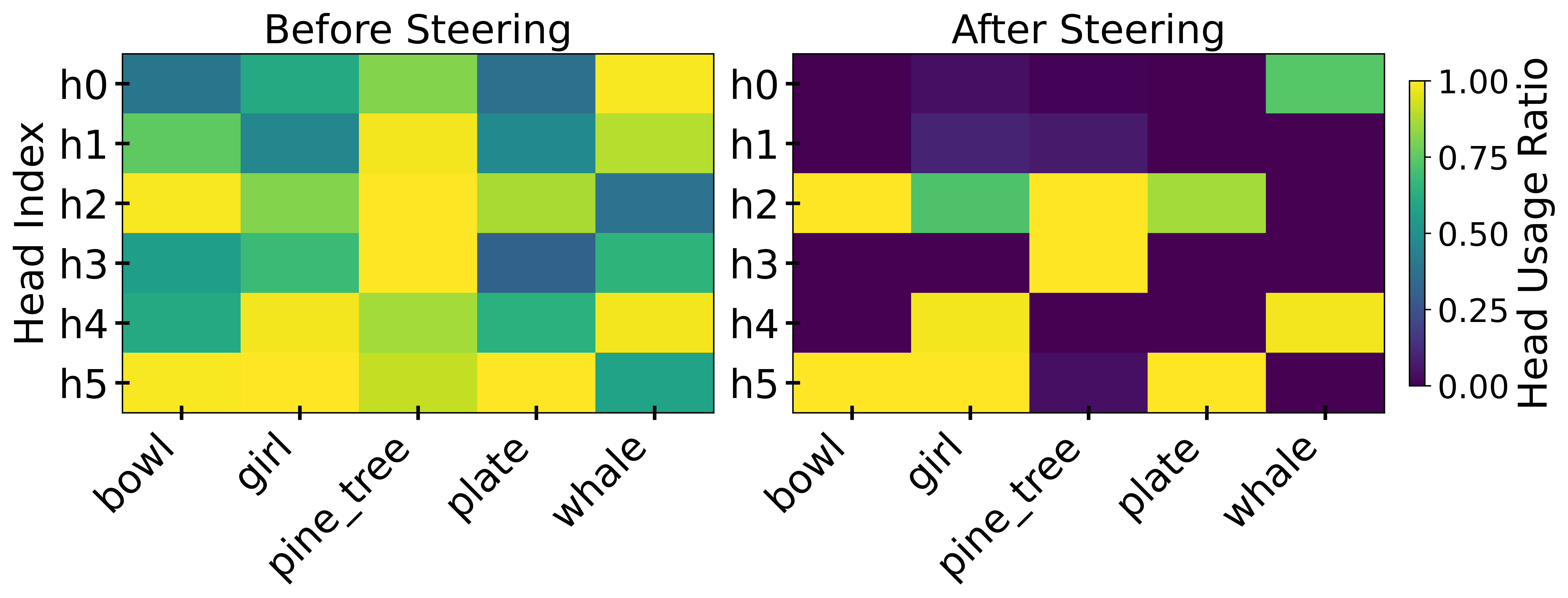}
    \caption{Effect of per-class steering ($\alpha = 1.2$) on head activation patterns. 
    Columns indicate the top-$5$ classes ranked by accuracy gain, 
    and rows correspond to heads ($h_0$–$h_5$).}
    \label{fig3}
\end{figure}

\paragraph{Steering dynamic pruning.}
Inspired by the top-$k$ masking experiments in PatchSAE \cite{lim2025patchsae}, 
we adapt this idea to dynamic head pruning using the latent steering defined in Eq.~(5), 
which amplifies selected SAE latent dimensions. 
For each sample, we evaluate three strategies for defining the index set $S$ based on training-set activation frequency statistics:
(1) \textbf{Per-class frequent} — top-$k$ most frequently activated latents within each class,  
(2) \textbf{Global frequent} — top-$k$ most frequently activated latents across all classes, and 
(3) \textbf{Random}.
Figure~\ref{fig2} shows that per-class steering reduces head usage while largely preserving accuracy, 
whereas global and random strategies lead to larger accuracy drops as $\alpha$ increases. 
The low overlap between global and per-class top-$k$ frequent latent dimensions ($0.1641$) indicates that the SAE captures class-discriminative concepts.
Figure~\ref{fig3} further illustrates head usage patterns in the final layer under per-class steering. 
For \textit{bowl}, accuracy rises ($76\% \!\to\! 82\%$) while head usage falls ($0.72 \!\to\! 0.33$), mainly relying on $h_2$ and $h_5$.
\textit{pine\_tree} shows a similar pattern ($79\% \!\to\! 84\%$, $0.93 \!\to\! 0.35$), relying on $h_2$ and $h_3$. 
Interestingly, semantically related classes such as \textit{bowl} and \textit{plate} share similar head subsets $h_2$ and $h_5$, indicating that per-class steering reveals class-level semantic relationships among heads.
These examples suggest that amplifying per-class top-$k$ frequent activations enriches class-specific signals in the decision network input, leading to class-specific pruning behaviors.
Overall, these results demonstrate that the Sparse Autoencoder provides an effective way to interpret and control dynamic head pruning in ViTs.

\section{Conclusion}
This paper introduces a novel Sparse Autoencoder-based framework that makes dynamic head pruning interpretable and controllable at the latent level in Vision Transformers.
By amplifying per-class frequent activations, we reveal class-specific pruning behaviors that are both efficient and interpretable.
Our current work focuses on the final layer and small datasets, and future work will extend the framework to earlier layers and foundation models.

\section{Acknowledgments}
This work was supported by the Technology Innovation Program (RS-2025-02613131) funded by the Ministry of Trade, Industry \& Energy (MOTIE, Korea).

\bibliography{aaai2026.bib}

\section{Appendix: Additional Analysis of SAE Latents}
We provide additional analysis of SAE latents, focusing on those extracted from the final-layer residual embedding, to study their role in controlling dynamic head pruning.
We examine (1) the effect of negative steering, (2) reconstruction behavior, and (3) the semantic structure of SAE latents.
These results indicate that SAE latents disentangle class-specific information into sparse dimensions, which are reflected in pruning decisions in the final layer.

\subsection{Effect of Negative Steering}
We observe that negative steering ($\alpha < 0$) increases head usage primarily for the per-class strategy, while global and random strategies show minimal change in head usage. In contrast, positive steering ($\alpha > 0$) reduces head usage. 
This suggests that negative steering suppresses class-specific latent features, weakening the pruning signal in the residual embedding, while positive steering amplifies them and leads to more selective head activation. 
As a result, this change modifies the input to the decision network, leading to different head activation patterns.

\subsection{Reconstruction Effect}
To isolate the effect of latent steering, we replace the original residual embedding with its SAE reconstruction.
Replacing the original residual embedding with its SAE reconstruction results in minimal changes in head usage (average $< 0.03$) and accuracy ($-0.12\%$), suggesting that SAE reconstruction preserves pruning-relevant information.
These results show that the observed effects are primarily due to latent steering rather than reconstruction artifacts.

\subsection{Semantic Structure of SAE Latents}
\begin{figure}[t]
    \centering
    \includegraphics[width=1.0\columnwidth]{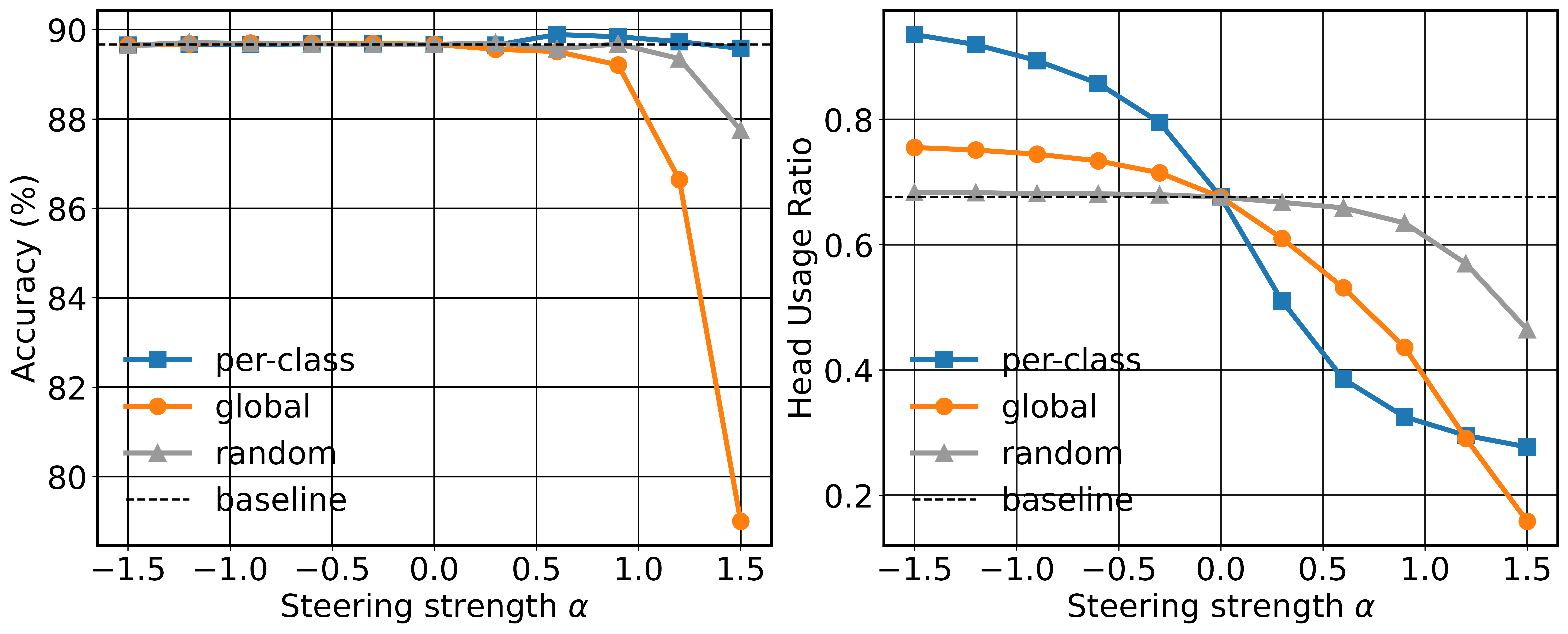}
    \caption{
    Negative steering ($\alpha < 0$) increases head usage by suppressing pruning signals.
    }
    \label{fig:neg_steering}
\end{figure}

\begin{figure}[t]
    \centering
    \includegraphics[width=0.69\columnwidth]{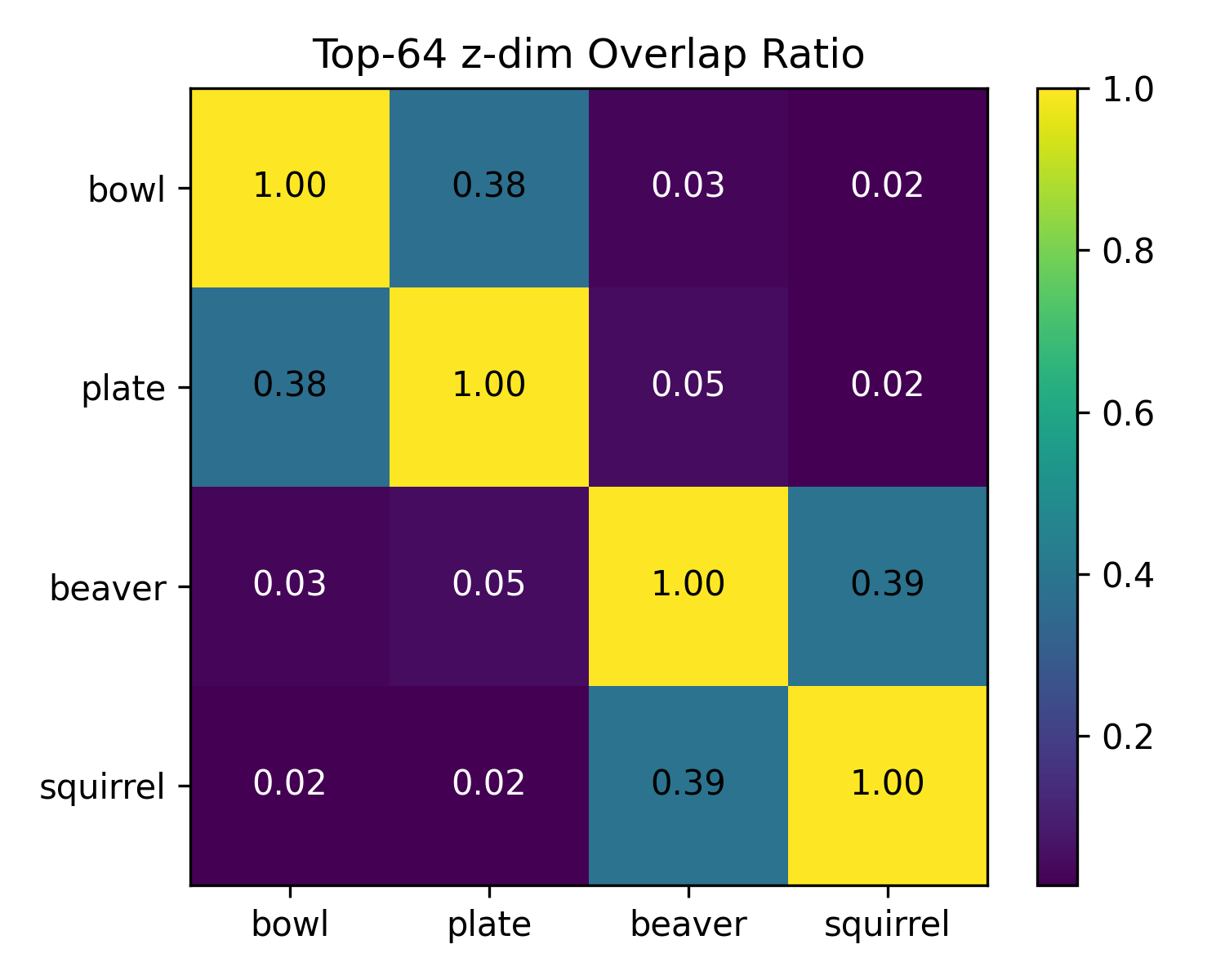}
    \includegraphics[width=1.0\columnwidth]{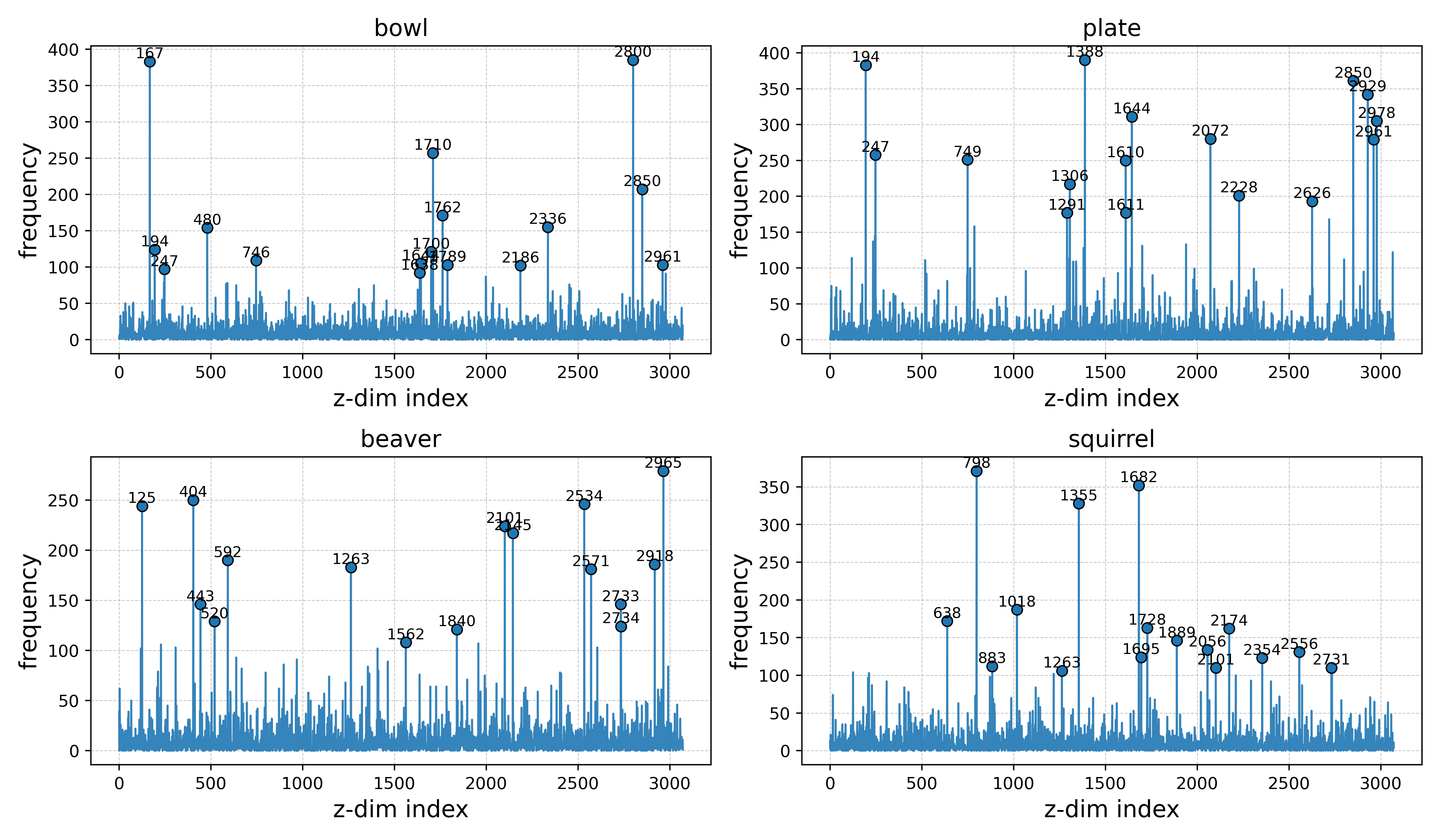}
    \caption{
    Semantic structure of SAE latents, showing class-wise overlap and distinct latent usage patterns.
    }
    \label{fig:latent_overlap}
\end{figure}

As shown in Figure~\ref{fig:latent_overlap}, semantically similar object classes such as \textit{bowl} and \textit{plate} exhibit high latent overlap ($\approx 0.38$), while unrelated classes show near-zero overlap. A similar pattern is observed for \textit{beaver} and \textit{squirrel} ($\approx 0.39$).
The figure also shows that each class relies on distinct sparse latent dimensions. 
These results demonstrate that SAE latents encode class-specific semantic structure, which is reflected in pruning decisions in the final layer.

\end{document}